\documentclass[conference]{IEEEtran}
\IEEEoverridecommandlockouts

\usepackage[T1]{fontenc}
\usepackage{graphicx}
\usepackage{fontspec}
\usepackage{enumitem}
\usepackage{multirow}
\usepackage{algorithm}
\usepackage{algpseudocode}
\usepackage{cite}
\usepackage{orcidlink}
\usepackage{array}
\usepackage{booktabs}
\usepackage{hyperref}
\hypersetup{
    colorlinks=false,
    pdfborder={0 0 0}
}
\usepackage{polyglossia}
\setdefaultlanguage{english}
\setotherlanguage{bengali}

\newfontfamily\bengalifont[
  Script=Bengali,
  Path= ./ ,
  Extension=.ttf
]{Kalpurush}

\usepackage{amsmath,amssymb,amsfonts}
\usepackage{textcomp}
\usepackage{xcolor}
\usepackage{tikz}
\usepackage{subcaption}
\usepackage{float}
\usepackage{csquotes}
\usepackage{siunitx}
\usepackage{tabularx}
\usepackage{threeparttable}
\usepackage{multicol}
\usepackage{xcolor,colortbl}
\usepackage{caption}
\usepackage[table]{xcolor}
\usepackage{makecell}

\definecolor{declight}{RGB}{235,235,235}   
\definecolor{intlight}{RGB}{255,240,210}   
\definecolor{implight}{RGB}{255,250,210}
\definecolor{excllight}{RGB}{255,225,210}

\usetikzlibrary{shapes, arrows, positioning, calc}

\usetikzlibrary{shapes,arrows}
\usepackage{verbatim}

\usepackage{url}
\usepackage{amsmath}
\usepackage{textcomp}
\usepackage{siunitx}
\usepackage{upgreek}

\makeatletter
 \let\old@ps@headings\ps@headings
 \let\old@ps@IEEEtitlepagestyle\ps@IEEEtitlepagestyle
 \def\confheader#1{%

 \def\ps@IEEEtitlepagestyle{%
 \old@ps@IEEEtitlepagestyle%
 \def\@oddhead{\strut#1\hfill\strut}%
 \def\@evenhead{\strut\hfill#1\hfill\strut}%
 }%
 \ps@headings%
 }
 \makeatother

\begin{document}
\title{Bangla Sentence Function Classification: Corpus Development, Model Benchmarking, and Interpretability}

\author{\IEEEauthorblockN{Swapnil Kundu Argha\IEEEauthorrefmark{1}, Abdullah Al Shafi\IEEEauthorrefmark{2}, Rowzatul Zannat\IEEEauthorrefmark{3}, Shoumik Barman Polok\IEEEauthorrefmark{4}, 
\\ Abdul Muntakim\IEEEauthorrefmark{5}, Jannatul Ferdousi\IEEEauthorrefmark{6}, M.A. Moyeen\IEEEauthorrefmark{7}}
\IEEEauthorblockA{
Department of CSE, Khulna University of Engineering \& Technology, Bangladesh \IEEEauthorrefmark{1} \IEEEauthorrefmark{2} \IEEEauthorrefmark{4} \IEEEauthorrefmark{7}\\
Institute of ICT, Khulna University of Engineering \& Technology, Bangladesh \IEEEauthorrefmark{2} \IEEEauthorrefmark{7} \\
Department of CSE, Daffodil International University, Bangladesh \IEEEauthorrefmark{3} \\
Department of Computer Science, Kennesaw State University, United States \IEEEauthorrefmark{5} \IEEEauthorrefmark{6}\\
swapnilkundu01@gmail.com\IEEEauthorrefmark{1}, abdullah@iict.kuet.ac.bd\IEEEauthorrefmark{2}, w.rzrowza@gmail.com\IEEEauthorrefmark{3}, polokbarman874@gmail.com\IEEEauthorrefmark{4}, 
\\ amuntaki@students.kennesaw.edu\IEEEauthorrefmark{5}, jannatulferdousi3110@gmail.com\IEEEauthorrefmark{6}, moyeen.kuet@gmail.com\IEEEauthorrefmark{7}}}


\maketitle
\begin{abstract}
Automatic sentence function identification is important for many downstream natural language processing (NLP) applications such as dialogue systems, text-to-speech synthesis, and machine translation. However, benchmark resources for Bangla sentence function classification remain limited. To mitigate this gap, this paper introduces a corpus of 10,000 Bangla sentences, manually annotated into four functional categories, namely declarative, interrogative, imperative, and exclamatory. The corpus is nearly balanced across the four classes, with high annotation reliability reflected by a Fleiss’ Kappa of 0.82. Furthermore, we evaluate multiple feature representations, including Bag-of-Words (BoW), TF-IDF, and Word2Vec, with several classical machine learning classifiers. In addition, two heterogeneous ensemble models, namely Single-Level Ensemble (SLE) and Double-Level Ensemble (DLE), are utilized to improve classification performance. Experimental results show that TF-IDF consistently outperforms Word2Vec, likely due to its ability to emphasize discriminative lexical cues associated with sentence functions, particularly given the relatively small corpus used to train Word2Vec. The DLE model with TF-IDF features achieves the best performance with accuracy and macro-F1 of 0.95, demonstrating the effectiveness of sparse lexical representations and heterogeneous ensemble learning for this task. Further cross-validation confirms the robustness of the approach, while LIME-based interpretability provides insights into model predictions. The developed corpus and model benchmarking establish strong baselines for Bangla sentence function classification.
\end{abstract}

\section{Introduction}
\label{sec:introduction}
With about 250 million native speakers, mostly in Bangladesh and West Bengal of India, Bangla is one of the most frequently spoken languages in the world. Compared to languages like English or Chinese, Bangla is still relatively deficient in automated language processing resources \cite{bijoy2021automated}. But automated systems require solutions to understand the functional expression of large amount of textual data as digital Bangla content has rapidly expanded its presence through various platforms \cite{al2026bist}.


Identifying sentence functions requires an evaluation of whether a sentence functions as declarative, interrogative, imperative, or exclamatory. The different functional categories allow a speaker to deliver his intended expression while controlling how listeners will understand the information delivered. For example, the declarative {\bengalifont{“সে বাড়িতে যায় (He goes home)”}} expresses a statement, while the interrogative {\bengalifont{“ভিতরে আসতে পারি? (May I come in?)”}} functions to obtain permission. Similarly, the imperative {\bengalifont{“গোলমাল করিও না। (Don't make a fuss.)”}} delivers a command, while the exclamatory {\bengalifont{“শৈশবের স্মৃতি কত মধুর! (How sweet childhood memories are!)”}} expresses strong feelings. 

Reliable functional classification of a sentence has broad implications across various downstream natural language processing (NLP) applications. Dialogue systems \cite{yi2025survey} rely on the identification of the type of query to generate appropriate responses. Machine translation \cite{castilho2025survey} benefits from preserving the intent of the sentence in multiple languages to maintain communicative accuracy. Sentiment and emotion detection can benefit from recognizing exclamatory and other functional cues, while stance detection can use the sentence function to better capture the communicative intent of the speaker \cite{alshafi-etal-2026-kuet}. Text-to-speech synthesis uses functional cues to produce suitable intonation patterns \cite{tan2024naturalspeech}. In short, without accurate understanding of sentence-level communicative function, higher-level NLP systems may fail to capture speaker intent, leading to reduced effectiveness in real-world language understanding.

The main contributions of this work are threefold.
\begin{itemize}
    \item We present a novel Bangla sentence function corpus with 10,000 manually annotated sentences across four functional categories. To ensure high-quality labeling of the corpus, a reliable multi-annotator annotation framework with agreement validation is designed. Furthermore, the corpus is released as a publicly available resource\footnote{\url{https://github.com/AbdullahRatulk/Bangla\_Sentence\_Function\_Classification\_Corpus}} to support reproducible research.
    \item A comprehensive benchmarking study for sentence function classification is conducted on various feature representations and classical machine learning models. We also benchmark two heterogeneous ensemble architectures to effectively combine diverse classifiers.
    \item We further incorporate interpretability and error analysis, providing insights into model behavior and linguistic challenges.
\end{itemize}

\section{Literature Review}
\label{sec:literature}
Early Bangla text classification research relied mainly on traditional machine learning methods due to the lack of large-scale datasets. Bijoy et al. \cite{bijoy2021automated} used ensemble models such as Random Forest and XGBoost for sentence classification, achieving 96.39\% accuracy, while Das et al. \cite{das2022analysis} applied Decision Tree models for structural sentence classification with 93.72\% accuracy. Furthermore, \cite{islam2018bard} introduced a large Bangla news dataset and showed that Word2Vec embeddings outperformed TF-IDF-based approaches, achieving a 0.96 F1-Score. 

With the rise of deep learning, Sikder et al. \cite{sikder2024hybrid} proposed a hybrid CNN-BiLSTM model that achieved 98\% F1-Score, demonstrating the effectiveness of neural architectures over traditional methods. Recently, Al Shafi et al. \cite{al2026bist} introduced BiST, a Bangla-English bilingual corpus for sentence structure and tense classification, with experiments showing the superiority of dual-encoder models over multilingual baselines.

\section{Proposed Method for Sentence Function Classification}
\label{sec:methodology}
\begin{figure}[htbp]
  \centering
  \makebox[0pt][c]{\includegraphics[width=0.48\textwidth]{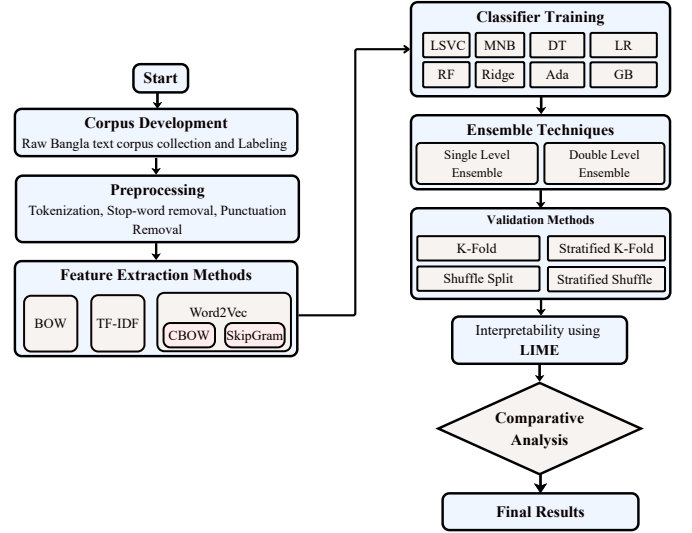}}%
  \caption{Overview of the proposed system.}
  \label{fig:proposed_methodology}
\end{figure}

The complete workflow of our proposed method is shown in Fig. \ref{fig:proposed_methodology}, which includes the sequence of corpus creation, data preprocessing, feature extraction, model training, evaluation, and interpretation.


\subsection{Development of Bangla Sentence Function Corpus}
\label{subsec:dataset}
\begin{figure}[htbp]
  \centering
  \includegraphics[width=0.48\textwidth]{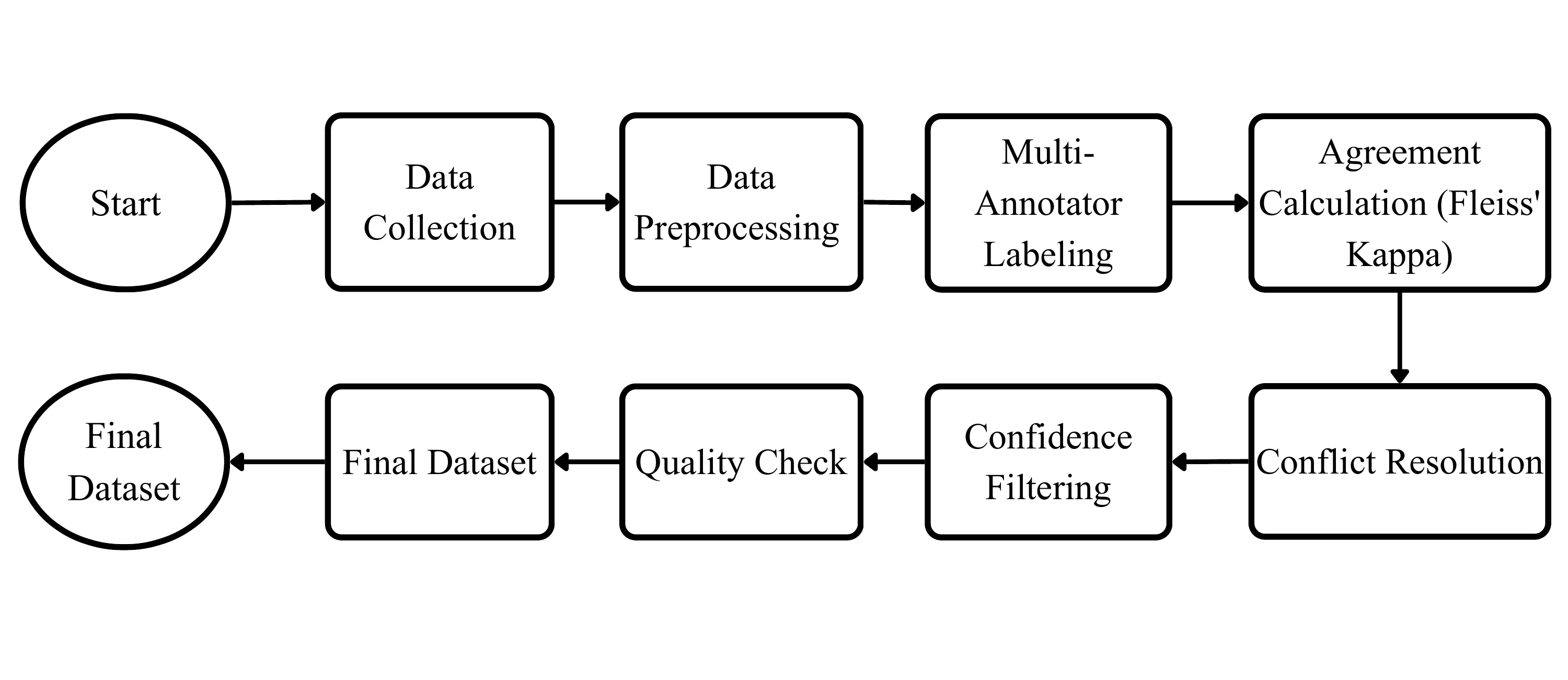}
  \caption{Workflow of the corpus development process.}
  \label{fig:data_collection_sample}
\end{figure}

The construction of a high-quality corpus is essential for robust text classification. Fig. \ref{fig:data_collection_sample} presents the workflow of the corpus creation process, including data collection, preprocessing, annotation, and quality verification.


\textbf{Data Collection and Preprocessing:} We prioritized collecting Bangla sentences that demonstrate different usage patterns through formal and informal usage across various online platforms which users can access without restrictions. Several popular Bengali newspapers serve as our primary sources that deliver high-quality formal content. To create a wider linguistic range, we also added Bengali blogs and online magazines and youtube comments together with digitized storybooks and novels which exist in the public domain. We have used automated web scraping techniques through Python scripts that used Beautiful Soup\footnote{https://pypi.org/project/beautifulsoup4/} to collect the initial raw sentences.


The collected text then underwent a rigorous preprocessing pipeline that includes removing HTML tags and non-Bengali characters, sentence segmentation using a Bengali sentence tokenizer, deduplication, and noise filtering to maintain focus on well-formed sentences.

\textbf{Annotation Guidelines:} Based on standard linguistic theory, we define four mutually exclusive functional classes, namely declarative, interrogative, imperative, and exclamatory for annotators.


A sentence was annotated as declarative if it conveyed a statement, fact, or opinion without expressing a query, command, or strong emotion. For example, \textquote*{\textbengali{আমি আজ স্কুলে গিয়েছিলাম।} (I went to school today.)}. In contrast, sentences were annotated as interrogative when they were intended to seek information or clarification, often characterized by the presence of interrogative markers such as \textbengali{কি, কেন, কোথায়, কবে, and কিভাবে}. For instance, \textquote{\textbengali{তুমি কোথায় যাচ্ছ?} (Where are you going?)}.

Sentences expressing commands, requests, suggestions, or instructions were categorized as imperative. \textquote*{\textbengali{দরজা বন্ধ করো।} (Close the door.)} would be an example of such a sentence. Meanwhile, sentences conveying strong emotions, such as surprise, joy, anger, or admiration, were labeled exclamatory. An example would be \textquote*{\textbengali{আহা! কী সুন্দর দৃশ্য!} (What a beautiful view!)}.

\begin{algorithm}
\caption{Corpus Annotation with Fleiss' Kappa}
\label{alg:annotation}
\begin{algorithmic}[1]
\Require Sentences $S=\{s_1,\dots,s_n\}$
\Ensure Final labels $F=\{f_1,\dots,f_n\}$
\State Define annotators $P=\{a_1,a_2,a_3\}$ and label sets $\mathcal{F}$
\ForAll{$s_i$ and $a_j$}
    \State Annotator assigns $f_{ij}\in\mathcal{F}$,
\EndFor
\ForAll{$s_i$}
    \State Compute item agreement $A_i$
\EndFor
\State $\bar{A}=\frac{1}{n}\sum_{i=1}^{n} A_i$
\State Compute expected agreement $A_e$
\State Fleiss' Kappa, $\kappa=\frac{\bar{A}-A_e}{1-A_e}$
\ForAll{$s_i$}
    \State $f_i \gets \underset{f \in \mathcal{F}}{\operatorname{arg\,max}} \sum_{j=1}^{3} \mathbf{1}(f_{ij} = f)$
    \If{no clear majority emerges}
        \State consensus discussion among annotators
    \EndIf
\EndFor

\Return Annotated corpus with final labels $F$
\end{algorithmic}
\end{algorithm}

\textbf{Corpus Annotation:} The multi-annotator framework which describes the corpus annotation process operates according to the steps which Algorithm \ref{alg:annotation} specifies. Three annotators assess each sentence in the corpus by applying their designated function labels from the defined function label set in annotation guideline. The reliability of the annotations is evaluated through inter-annotator agreement (IAA) assessment which uses Fleiss' Kappa method. The final label for each sentence comes using majority voting. The final label assignment process requires annotators to hold a consensus discussion when a situation arises without a distinct majority. The resulting labels collectively form the finalized annotated corpus.

\textbf{Final Corpus:} The final corpus was created from the original collection of approximately 24,000 Bangla sentences from which 10,000 sentences underwent preprocessing and annotation to establish a reliable benchmark for sentence function classification.

\subsection{Data Preprocessing}
\label{subsec:preprocessing_pipeline}
The raw Bangla sentences undergo a series of cleaning and preprocessing steps before feature extraction to reduce noise while preserving linguistic cues essential for function classification. The cleaning phase applies sequential operations, including the elimination of hashtags and special characters that do not carry linguistic meaning, space normalization, etc. Common stop words for Bangla\footnote{https://github.com/stopwords-iso/stopwords-bn} are removed with a critical exception made for wh-question words such as {\bengalifont{কী}} (what), {\bengalifont{কেমন}} (how), {\bengalifont{কখন}} (when), {\bengalifont{কোথায়}} (where), as these words are essential for identifying interrogative and/or exclamatory sentences. All punctuation marks are removed to ensure that punctuation does not serve as a direct cue for sentence function, forcing the classification model to rely on lexical and structural cues instead. A Bangla-aware custom tokenizer splits text on whitespace and returns only non-empty tokens after all cleaning steps, with the same tokenizer used across all feature extractors for fair comparison. In addition, the four function labels are encoded into integer IDs from 0 to 3 for model training.


\subsection{Feature Extraction}
\label{subsec:feature_extraction}
We employ multiple feature extraction techniques to represent the preprocessed text as numerical vectors, namely Bag of Words (BoW), TF-IDF, and Word2Vec.

BoW \cite{das2022analysis} counts the frequency of each word or n-gram in a document while ignoring word order and grammar, implemented with CountVectorizer using configurable n-gram ranges. On the other hand, TF-IDF \cite{bijoy2021automated} weights terms by their frequency in a document and their rarity across the corpus according to the mathematical formulation where TF represents term frequency in a document and IDF represents inverse document frequency across the corpus. For example, the term {\bengalifont{কী}} (what) appearing in an interrogative document receives high TF in that document and moderate IDF, resulting in high TF-IDF weight that marks it as a useful feature for identifying interrogatives.

\begin{figure}[htbp]
  \centering
  \makebox[0pt][c]{\includegraphics[width=0.54\textwidth]{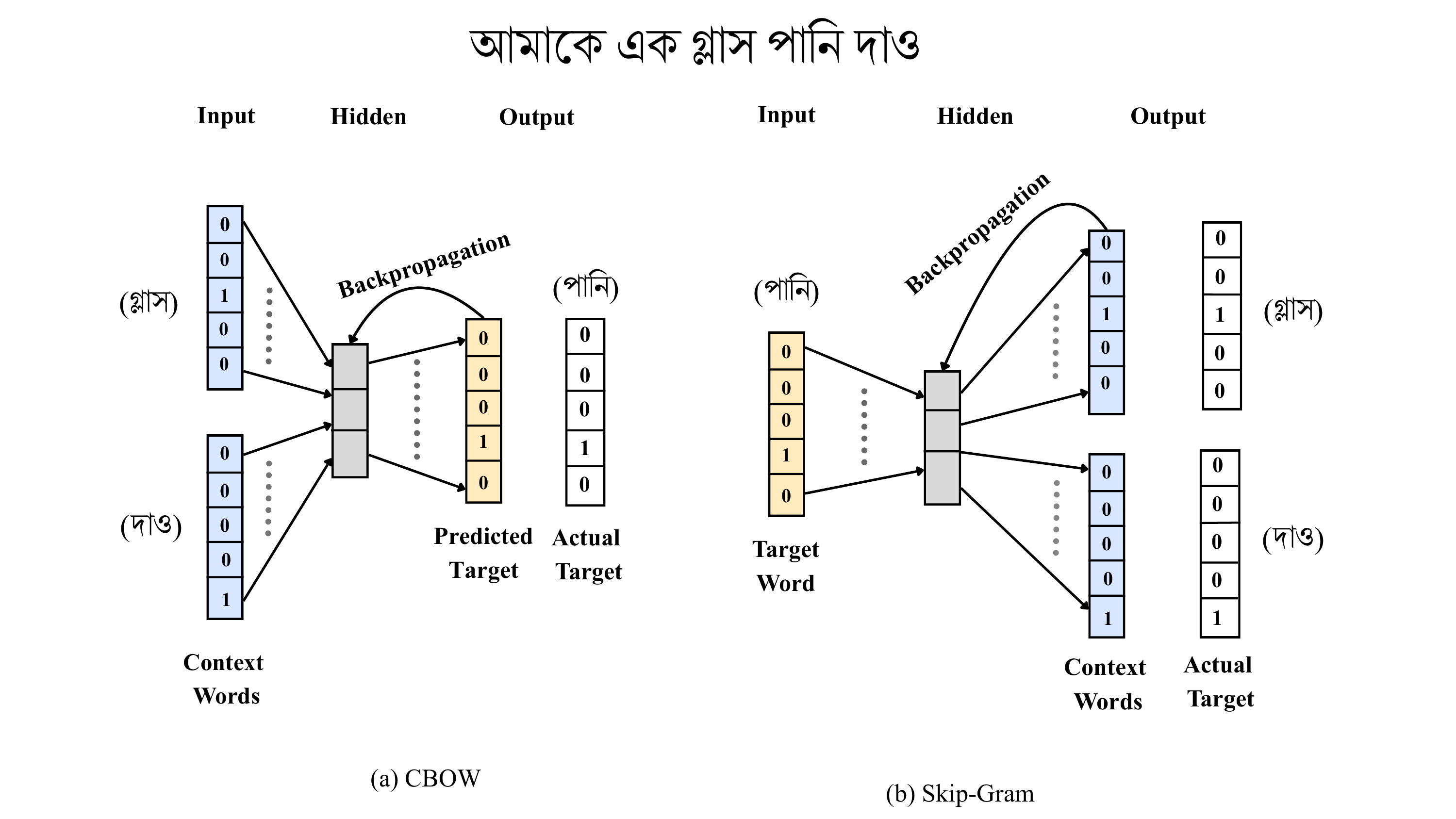}}%
  \caption{Comparison of (a) CBOW and (b) Skip-Gram architectures where CBOW predicts the target word from surrounding context words while Skip-Gram predicts context words from the target word.}
  \label{fig:cbow_skipgram}
\end{figure}

\begin{figure*}
  \centering
  \begin{subfigure}[t]{0.44\textwidth}
    \centering
    \includegraphics[width=\textwidth]{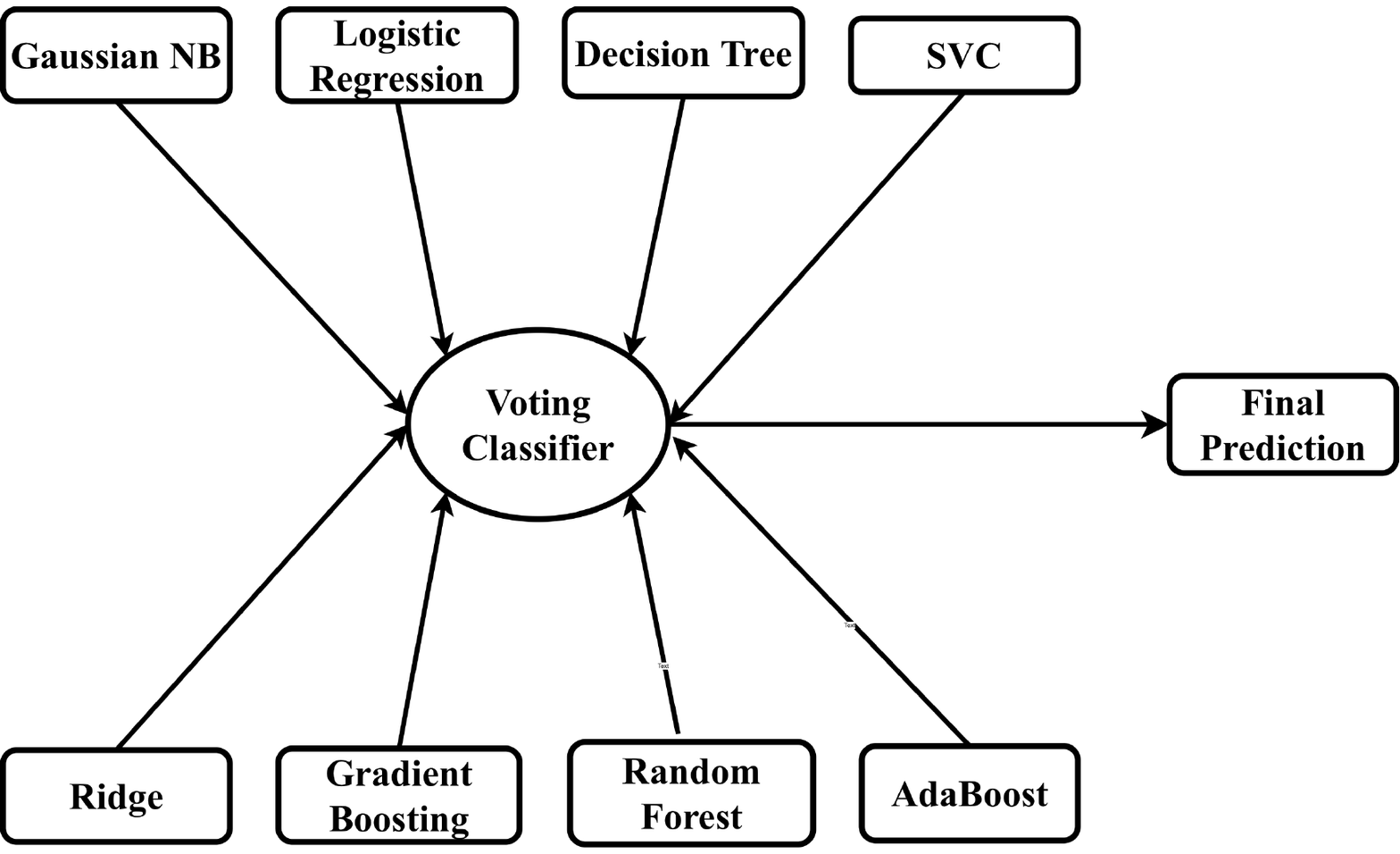}
    \caption{\hspace{0.3em}Single-Level Ensemble (SLE) architecture}   
    \label{fig:sle_arch}
  \end{subfigure}
  \vspace{5pt}
  \begin{subfigure}[t]{0.44\textwidth}
    \centering
    \includegraphics[width=\textwidth]{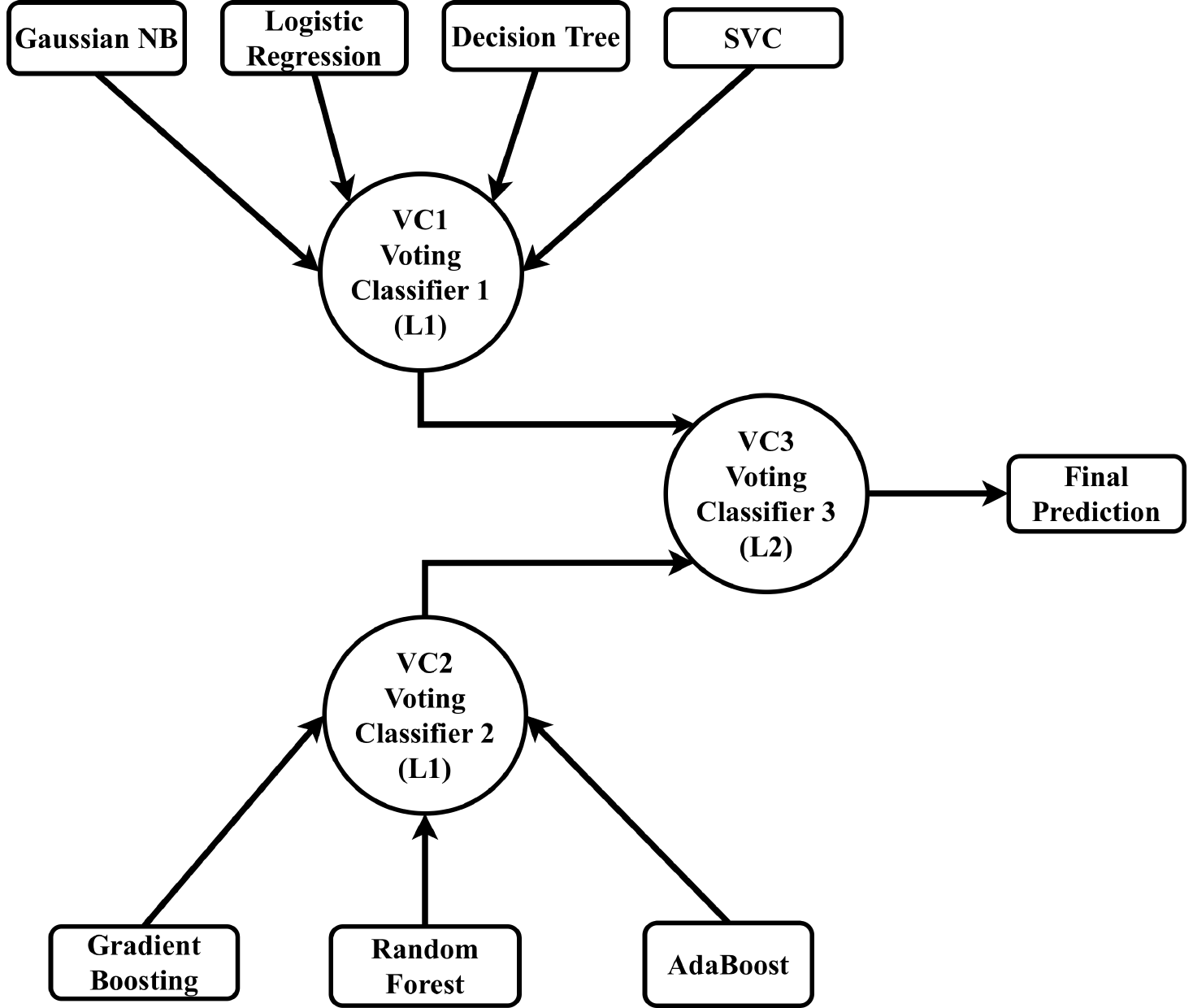}
    \caption{\hspace{0.3em}Double-Level Ensemble (DLE) architecture}   
    \label{fig:dle_arch}
  \end{subfigure}
  \caption{Architecture of the two heterogeneous ensemble models: (a) SLE where multiple classical classifiers feed into a Voting Classifier for final prediction; (b) DLE with two levels of voting classifiers leading to final prediction.}
  \label{fig:sle_dle_architecture}
\end{figure*}

In contrast, Word2Vec \cite{islam2018bard} is a two-layer neural network trained on the corpus to capture semantic relationships using two different architectures as shown in Fig. \ref{fig:cbow_skipgram}: (i) Continuous Bag of Words (CBOW) which predicts the target word from its surrounding context and performs well for frequent words, and (ii) Skip-gram which predicts context words from the target word and tends to capture rare words and semantic relationships better. 

\subsection{Classification} 
We evaluate multiple classifiers spanning classical ML and heterogeneous ensemble approaches.


\textbf{Classical ML Classifiers: }The classical classifiers trained using the extracted features include Decision Tree (DT), Multinomial Naive Bayes (MNB), Logistic Regression (LR), Linear Support Vector Classifier (LSVC), Ridge classifier, Gradient Boosting (GB), AdaBoost (Ada), and Random Forest (RF).

\textbf{Heterogeneous Ensemble Classifiers: }We design two heterogeneous ensemble architectures, namely Single Level Ensemble (SLE) and Double Level Ensemble (DLE), to leverage the complementary strengths of different classical classification algorithms. Fig. \ref{fig:sle_dle_architecture} presents a detailed overview of both architectures.



SLE encompasses all classical ML classifiers operating on fixed feature vectors. These classifiers then feed into a voting classifier to produce the final classification output. On the other hand, DLE implements a more sophisticated hierarchical voting mechanism with two distinct aggregation levels. In the first level, the base classifiers are organized into two groups based on their underlying algorithmic characteristics. Group 1 consists of MNB, LR, DT, and LSVC, whose outputs are combined into voting classifier 1 (VC1). Group 2 comprises the homogeneous ensemble methods including RF, GB, and AdaBoost, whose predictions are aggregated into voting classifier 2 (VC2). In the second level, the outputs from VC1 and VC2 are fed into a third voting classifier (VC3) that performs the final aggregation to produce the ultimate prediction. This hierarchical structure with two layers of voting enables the model to capture different levels of decision making, first within similar classifier types and then across the two broad categories of classifiers.

It is to be noted that the voting classifiers use soft voting, where the final prediction is obtained by averaging the predicted probabilities of all base classifiers and selecting the class with the highest score.

\textbf{Cross-Validation and Hyperparameter Tuning:} Robust evaluation is ensured through multiple cross-validation (CV) strategies including (i) K-fold where data is split into K folds with the model training on (K-1) folds and validating on 1 fold rotating K times, (ii) stratified K-fold which maintains original class proportions in each fold, (iii) shuffle split with multiple random train-validation splits, and (iv) stratified shuffle split combining random splitting with stratification. Grid search with 5-fold CV is used to optimize hyperparameters for each model with macro-F1 score as the objective metric, and after identifying the best parameters, models are refitted on the full training set.


\subsection{Interpretability using LIME}
The research employs Local Interpretable Model-agnostic Explanations (LIME) \cite{salih2025perspective} to analyze the model's prediction results. This local approximation method enables us to identify which words or phrases have the greatest impact on model's prediction.

\section{Experimental setup and Result analysis}
\label{sec:experiments}
\begin{table}[htbp]
\centering
\caption{Class distribution and sentence length statistics for the Bangla Sentence Function Corpus.}
\label{tab:class_distribution}
\begin{tabular}{lccccc}
\toprule
\textbf{Function} & \textbf{Total} & \multicolumn{3}{c}{\textbf{Length Bins (Tokens)}} & \textbf{Avg. Len} \\
\cmidrule(lr){3-5}
 & & 0-8 & 9-14 & 15+ & \\
\midrule
Declarative   & 2,487 & 881 & 932 & 674 & 10.78 \\
Interrogative & 2,532 & 895 & 902 & 735 & 11.02 \\
Imperative    & 2,511 & 845 & 833 & 833 & 11.29 \\
Exclamatory   & 2,470 & 804 & 833 & 833 & 12.52 \\
\midrule
Total & 10,000 & 3,425 & 3,500 & 3,075 & 11.40\\
\bottomrule
\end{tabular}
\end{table}

\begin{table*}[t]
\centering
\caption{Comparative analysis of various feature representation and classifiers based on performance evaluation. All the metrics except accuracy are macro-averaged.}
\label{tab:comparative_full}
\footnotesize
\renewcommand{\arraystretch}{1.1}
\setlength{\tabcolsep}{3pt}
\begin{tabular}{llcccccccccc}
\hline
\textbf{Metric} & \textbf{Feature} & \textbf{NB} & \textbf{LSVC} & \textbf{DT} & \textbf{LR} & \textbf{RF} & \textbf{Ridge} & \textbf{Ada} & \textbf{GB} & \textbf{SLE} & \textbf{DLE} \\
\hline
\multirow{4}{*}{\textbf{Accuracy}}
 & BOW      & 0.86 & 0.93 & 0.84 & 0.93 & 0.94 & 0.90 & 0.88 & 0.94 & 0.94 & \textbf{0.94} \\
 & TF-IDF   & 0.87 & 0.94 & 0.85 & 0.94 & 0.94 & 0.92 & 0.89 & 0.94 & 0.94 & \textbf{0.95} \\
 & CBOW     & 0.83 & 0.90 & 0.81 & 0.89 & 0.91 & 0.89 & 0.86 & 0.91 & \textbf{0.93} & 0.92 \\
 & SkipGram & 0.84 & 0.91 & 0.82 & 0.90 & 0.92 & 0.90 & 0.87 & 0.92 & 0.93 & \textbf{0.94} \\
\hline
\multirow{4}{*}{\textbf{Precision}}
 & BOW      & 0.86 & 0.93 & 0.84 & 0.94 & 0.94 & 0.91 & 0.88 & 0.94 & 0.94 & \textbf{0.95} \\
 & TF-IDF   & 0.87 & 0.94 & 0.85 & 0.94 & 0.94 & 0.92 & 0.89 & 0.94 & 0.94 & \textbf{0.95} \\
 & CBOW     & 0.83 & 0.90 & 0.81 & 0.89 & 0.91 & 0.89 & 0.86 & 0.91 & \textbf{0.94} & 0.93 \\
 & SkipGram & 0.84 & 0.91 & 0.82 & 0.90 & 0.92 & 0.90 & 0.87 & 0.92 & 0.93 & \textbf{0.94} \\
\hline
\multirow{4}{*}{\textbf{Recall}}
 & BOW      & 0.86 & 0.93 & 0.84 & 0.93 & 0.94 & 0.90 & 0.88 & 0.94 & 0.94 & \textbf{0.95} \\
 & TF-IDF   & 0.87 & 0.94 & 0.85 & 0.94 & 0.94 & 0.92 & 0.89 & 0.94 & 0.94 & \textbf{0.95} \\
 & CBOW     & 0.83 & 0.90 & 0.81 & 0.89 & 0.91 & 0.89 & 0.86 & 0.91 & \textbf{0.93} & \textbf{0.93} \\
 & SkipGram & 0.84 & 0.91 & 0.82 & 0.90 & 0.92 & 0.90 & 0.87 & 0.92 & 0.93 & \textbf{0.94} \\
\hline
\multirow{4}{*}{\textbf{F1-Score}}
 & BOW      & 0.86 & 0.93 & 0.84 & 0.93 & 0.94 & 0.90 & 0.88 & 0.94 & 0.94 & \textbf{0.94} \\
 & TF-IDF   & 0.87 & 0.94 & 0.85 & 0.94 & 0.94 & 0.92 & 0.89 & 0.94 & 0.94 & \textbf{0.95} \\
 & CBOW     & 0.83 & 0.90 & 0.81 & 0.89 & 0.91 & 0.89 & 0.86 & 0.91 & \textbf{0.94} & 0.93 \\
 & SkipGram & 0.84 & 0.91 & 0.82 & 0.90 & 0.92 & 0.90 & 0.87 & 0.92 & 0.93 & \textbf{0.94} \\
\hline
\end{tabular}
\end{table*}

\begin{table*}[htbp]
\caption{Cross-validation performance evaluation of SLE and DLE models (mean ± std). Here, we set K = 5.}
\label{tab:cv}
\centering
\footnotesize
\setlength{\tabcolsep}{3pt}
\begin{tabular}{llcc|cc}
\toprule
\multirow{2}{*}{\textbf{CV techniques}} & \multirow{2}{*}{\textbf{Features}} & \multicolumn{2}{c}{\textbf{SLE}} & \multicolumn{2}{c}{\textbf{DLE}} \\
\cmidrule(lr){3-4} \cmidrule(lr){5-6}
 & & \textbf{Accuracy} & \textbf{Macro-F1} & \textbf{Accuracy} & \textbf{Macro-F1} \\
\midrule
\multirow{4}{*}{K-Fold}
 & BoW      & 0.94±0.012 & 0.94±0.012 & 0.95±0.009 & 0.95±0.009 \\
 & TF-IDF   & 0.94±0.011 & 0.94±0.011 & 0.95±0.008 & 0.95±0.008 \\
 & CBOW     & 0.92±0.018 & 0.92±0.018 & 0.93±0.015 & 0.93±0.015 \\
 & SkipGram & 0.93±0.015 & 0.93±0.015 & 0.94±0.012 & 0.94±0.012 \\
\midrule
\multirow{4}{*}{Stratified K-Fold}
 & BoW      & 0.94±0.010 & 0.94±0.010 & 0.95±0.007 & 0.95±0.007 \\
 & TF-IDF   & 0.94±0.009 & 0.94±0.009 & 0.95±0.006 & 0.95±0.006 \\
 & CBOW     & 0.92±0.015 & 0.92±0.015 & 0.93±0.012 & 0.93±0.012 \\
 & SkipGram & 0.93±0.012 & 0.93±0.012 & 0.94±0.010 & 0.94±0.010 \\
\midrule
\multirow{4}{*}{Shuffle Split}
 & BoW      & 0.94±0.018 & 0.94±0.018 & 0.95±0.014 & 0.95±0.014 \\
 & TF-IDF   & 0.94±0.016 & 0.94±0.016 & 0.95±0.012 & 0.95±0.012 \\
 & CBOW     & 0.92±0.024 & 0.92±0.024 & 0.93±0.020 & 0.93±0.020 \\
 & SkipGram & 0.93±0.020 & 0.93±0.020 & 0.94±0.016 & 0.94±0.016 \\
\midrule
\multirow{4}{*}{Stratified Shuffle Split}
 & BoW      & 0.94±0.014 & 0.94±0.014 & 0.95±0.011 & 0.95±0.011 \\
 & TF-IDF   & 0.94±0.013 & 0.94±0.013 & 0.95±0.010 & 0.95±0.010 \\
 & CBOW     & 0.92±0.020 & 0.92±0.020 & 0.93±0.017 & 0.93±0.017 \\
 & SkipGram & 0.93±0.017 & 0.93±0.017 & 0.94±0.014 & 0.94±0.014 \\
\bottomrule
\end{tabular}
\end{table*}

\begin{table*}[htbp]
\centering
\caption{Representative genuine misclassification cases with DLE on TF-IDF features.}
\label{tab:error_analysis}
\footnotesize
\setlength{\tabcolsep}{2pt}
\begin{tabular}{|p{7.7cm}|c|c|p{4.7cm}|}
\hline
\textbf{Text} & \textbf{Actual} & \textbf{Predicted} & \textbf{Primary Error Source} \\
\hline

{\bengalifont{বাংলাদেশের এই প্রাকৃতিক দৃশ্য কত মনোমুগ্ধকর!}} 
{\small (How mesmerizing this natural scenery of Bangladesh is!)} 
& \cellcolor{excllight} Exclamatory 
& \cellcolor{declight} Declarative 
& Exclamatory sentence with declarative syntax. \\
\hline

{\bengalifont{এই নীতিমালার কার্যকারিতা নিয়ে কোনো সন্দেহ আছে?}} 
{\small (Is there any doubt about the effectiveness of this policy?)} 
& \cellcolor{intlight} Interrogative 
& \cellcolor{declight} Declarative 
& Rhetorical question interpreted as statement. \\
\hline

{\bengalifont{সংশ্লিষ্ট কর্তৃপক্ষকে দ্রুত প্রয়োজনীয় ব্যবস্থা গ্রহণ করতে হবে।}} 
{\small (The relevant authorities must take necessary measures promptly.)} 
& \cellcolor{declight} Declarative 
& \cellcolor{implight} Imperative 
& Obligation modality interpreted as directive.\\
\hline

{\bengalifont{গবেষণা অনুযায়ী জলবায়ু পরিবর্তনের প্রভাব ক্রমেই বৃদ্ধি পাচ্ছে।}} 
{\small (According to research, the impact of climate change is increasing.)} 
& \cellcolor{declight} Declarative 
& \cellcolor{excllight} Exclamatory 
& Strong evaluative tone misinterpreted as exclamation. \\
\hline

\end{tabular}
\end{table*}

\subsection{Inter-Annotator Agreement}
The Fleiss' Kappa ($\kappa$) value for the developed corpus is 0.82, which is treated as \textquote*{Almost Perfect} in the standard, indicating a high level of inter-annotator agreement and strong consistency in the labeling process.


\subsection{Corpus Statistics}
\label{subsubsec:selection}
As shown in Table \ref{tab:class_distribution}, each of the four functional classes contributes approximately 2,500 sentences, ensuring a near-equal distribution that prevents bias towards any single function. The three sentence length analysis bins which include 0-8 tokens and 9-14 tokens and 15+ tokens show that all functional classes use a combination of brief sentences, intermediate sentences, and extended sentences to create a corpus through which we can train models that can handle different input lengths.

\subsection{Experimental Setup and Evaluation Metrics}
The final curated dataset was partitioned into two subsets using stratified sampling, a training-validation set of 8000 sentences, and an unseen test set of 2000 sentences for the final unbiased evaluation. Model performance is assessed using standard metrics, including accuracy, precision, recall, and F1-Score.



\subsection{Result Analysis}
Table \ref{tab:comparative_full} presents the comprehensive performance comparison across four feature representations and eight classifiers. Linear models, including LSVC, Ridge, and LR demonstrate strong performance across most feature types. RF and GB emerged as the top performers, RF particularly excelled with TF-IDF features reaching 0.94 accuracy and F1-Score. DLE achieved 0.95 on the TF-IDF features, while SLE maintained 0.94 on most metrics.

In addition, the superior performance of TF-IDF over Word2Vec can be attributed to the discriminative lexical markers of sentence functions, such as interrogative words, directive, and exclamatory expressions, etc. TF-IDF emphasizes such class-specific terms, whereas Word2Vec captures broader semantic relationships and may not preserve these functional cues as explicitly, particularly when trained on a relatively small corpus. Thus, sparse lexical weighting is well suited to this task.

The heterogeneous ensembles are also tested against several different CV methods to confirm their robustness, and the results are shown in Table \ref{tab:cv}. The results confirm the stability of our ensemble approaches. Both SLE and DLE reach 92-95\% accuracy with standard deviations between 0.6\% and 2.4\% for all feature representation.

Table \ref{tab:error_analysis} presents representative examples of such misclassified sentences which highlight common challenges. Exclamatory sentences are frequently misclassified as declarative when exclamation expressions are underweighted. The use of directive verbs creates problems that lead to difficulties in determining between declarative and imperative sentence structures. Questions that contain embedded commands are sometimes treated as direct orders. 

\begin{figure}[htbp]
\centering
\includegraphics[width=0.5\textwidth]{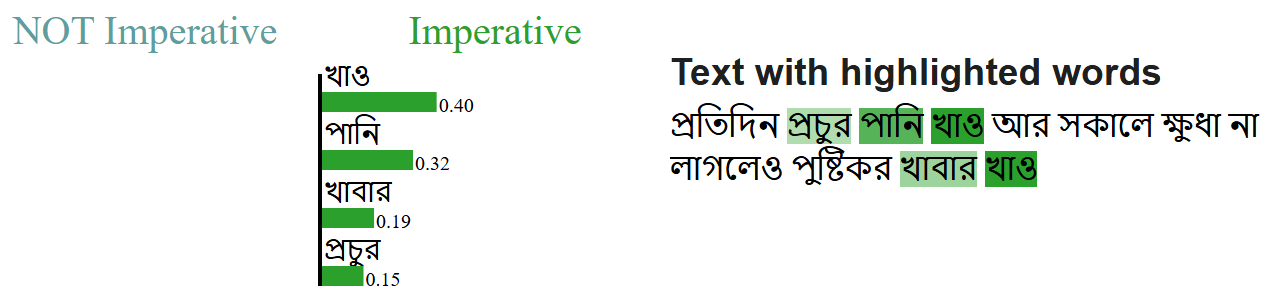}
\caption{Interpreting model decisions through LIME, visualizing feature importance for predicted sentence class.}
\label{fig:explainability}
\end{figure}


Fig. \ref{fig:explainability} provides an example of clear and intuitive illustration of the model’s decision-making process using LIME, highlighting the contribution of individual words toward the predicted Imperative class. Key token \textquote*{\textbengali{খাও}} is assigned higher importance scores, which aligns well with their semantic association with actions and directives. This indicates that the model is able to capture meaningful lexical cues relevant to imperative sentence construction. Similar patterns were observed for the other classes.


\section{Conclusion}
\label{sec:conclusion}
The research presents a high-quality Bangla sentence function corpus together with a study that compares traditional learning methods and heterogeneous ensemble learning techniques. The experimental findings demonstrate that TF-IDF which represents data in sparse form continues to perform effectively for the current task. The ensemble methods demonstrate superior performance stability among all tested models whereas DLE produces the highest overall performance results. The analysis of errors shows difficulties that arise from linguistic uncertainty which exists between different types of sentences. The newly developed corpus together with its benchmark tests offers valuable resources which enable future researchers to advance Bangla NLP studies. The future direction would be using transformer-based language models and larger datasets to enhance both classification accuracy and system resilience.

\bibliographystyle{./IEEEtran}
\bibliography{./IEEEexample}

@article{sikder2024hybrid,
  title={A hybrid approach for Bengali sentence validation},
  author={Sikder, Juel and Chakraborty, Prosenjit and Das, Utpol Kanti and Dhar, Krity},
  journal={Artificial Intelligence Review},
  volume={57},
  number={11},
  pages={316},
  year={2024},
  publisher={Springer}
}

@inproceedings{al2026bist,
  title={BiST: A Gold Standard Bangla-English Bilingual Corpus for Sentence Structure and Tense Classification with Inter-Annotator Agreement},
  author={Al Shafi, Abdullah and Argha, Swapnil Kundu and Moyeen, MA and Muntakim, Abdul and Polok, Shoumik Barman},
  booktitle={SIGUL 2026 Joint Workshop with ELE, EURALI, and DCLRL: Towards Inclusivity and Equality: Language Resources and Technologies for Under-Resourced and Endangered Languages@ LREC 2026},
  pages = {143--152},
  year={2026},
  publisher = {European Language Resources Association (ELRA)},
}

@inproceedings{alshafi-etal-2026-kuet,
  title = {StanceMoE: Mixture-of-Experts Architecture for Stance Detection},
  author = {Al Shafi, Abdullah and Islam, Md. Milon and Hossain, Sk. Imran and Hasan, K. M. Azharul},
  booktitle = {Proceedings of the 2nd International Workshop on Nakba Narratives as Language Resources @ LREC 2026},
  year = {2026},
  pages = {244--251},
  publisher = {European Language Resources Association (ELRA)},
  doi = {10.63317/3pim9v9kog3d}
}

@article{yi2025survey,
  title={A survey on recent advances in llm-based multi-turn dialogue systems},
  author={Yi, Zihao and Ouyang, Jiarui and Xu, Zhe and Liu, Yuwen and Liao, Tianhao and Luo, Haohao and Shen, Ying},
  journal={ACM Computing Surveys},
  volume={58},
  number={6},
  pages={1--38},
  year={2025},
  publisher={ACM New York, NY}
}

@article{castilho2025survey,
  title={A survey of context in neural machine translation and its evaluation},
  author={Castilho, Sheila and Knowles, Rebecca},
  journal={Natural Language Processing},
  volume={31},
  number={4},
  pages={986--1016},
  year={2025},
  publisher={Cambridge University Press}
}

@article{salih2025perspective,
  title={A perspective on explainable artificial intelligence methods: SHAP and LIME},
  author={Salih, Ahmed M and Raisi-Estabragh, Zahra and Galazzo, Ilaria Boscolo and Radeva, Petia and Petersen, Steffen E and Lekadir, Karim and Menegaz, Gloria},
  journal={Advanced Intelligent Systems},
  volume={7},
  number={1},
  pages={2400304},
  year={2025},
  publisher={Wiley Online Library}
}

@inproceedings{bijoy2021automated,
  title={An automated approach for Bangla sentence classification using supervised algorithms},
  author={Bijoy, Md Hasan Imam and Hasan, Mehedi and Tusher, Abdur Nur and Rahman, Md Mahbubur and Mia, Md Jueal and Rabbani, Masud},
  booktitle={2021 12th International Conference on Computing Communication and Networking Technologies (ICCCNT)},
  pages={1--6},
  year={2021},
  organization={IEEE}
}

@inproceedings{islam2018bard,
  author={Islam, Md Mofijul and Alam, Md. Tanvir},
  title={BARD: Bangla Article Classification Using a New Comprehensive Dataset},
  booktitle={2018 International Conference on Bangla Speech and Language Processing (ICBSLP)},
  year={2018}
}

@inproceedings{das2022analysis,
  title={Analysis of Bangla transformation of sentences using machine learning},
  author={Das, Rajesh Kumar and Sammi, Samrina Sarkar and Kobra, Khadijatul and Ajmain, Moshfiqur Rahman and khushbu, Sharun Akter and Noori, Sheak Rashed Haider},
  booktitle={International Conference on Deep Learning, Artificial Intelligence and Robotics},
  pages={36--52},
  year={2022},
  organization={Springer}
}

@article{tan2024naturalspeech,
  title={Naturalspeech: End-to-end text-to-speech synthesis with human-level quality},
  author={Tan, Xu and Chen, Jiawei and Liu, Haohe and Cong, Jian and Zhang, Chen and Liu, Yanqing and Wang, Xi and Leng, Yichong and Yi, Yuanhao and He, Lei and others},
  journal={IEEE Transactions on Pattern Analysis and Machine Intelligence},
  volume={46},
  number={6},
  pages={4234--4245},
  year={2024},
  publisher={IEEE}
}

\end{document}